\documentclass[runningheads,a4paper]{llncs}
\usepackage{graphicx}
\usepackage{amssymb}
\usepackage{subfigure}
\usepackage{array}
\usepackage{bm}
\usepackage{multirow}
\usepackage{cite}
\usepackage{verbatim}
\usepackage{mathrsfs}

\usepackage{url}
\urldef{\mailsa}\path|{alfred.hofmann, ursula.barth, ingrid.haas, frank.holzwarth,|
\urldef{\mailsb}\path|anna.kramer, leonie.kunz, christine.reiss, nicole.sator,|
\urldef{\mailsc}\path|erika.siebert-cole, peter.strasser, lncs}@springer.com|
\newcommand{\keywords}[1]{\par\addvspace\baselineskip
\noindent\keywordname\enspace\ignorespaces#1}

\begin{document}

\mainmatter  % start of an individual contribution

\titlerunning{HDSR with residual network and RNN-CTC}
\title{Handwritten digit string recognition by combination of residual network and RNN-CTC}

\author{Hongjian Zhan, Qingqing Wang, Yue Lu}% the name(s) of the author(s) follow(s) next

\institute{Shanghai Key Laboratory of Multidimensional Information Processing\\
Department of Computer Science and Technology\\
East China Normal University, Shanghai 200062, China\\
\url{ylu@cs.ecnu.edu.cn}}

\toctitle{Lecture Notes in Computer Science}
\tocauthor{Authors' Instructions}
\maketitle

\begin{abstract}
Recurrent neural network (RNN) and connectionist temporal classification (CTC) have showed successes in many sequence labeling tasks with the strong ability of dealing with the problems where the alignment between the inputs and the target labels is unknown. Residual network is a new structure of convolutional neural network and works well in various computer vision tasks. In this paper, we take advantage of the architectures mentioned above to create a new network for handwritten digit string recognition. First we design a residual network to extract features from input images, then we employ a RNN to model the contextual information within feature sequences and predict recognition results. At the top of this network, a standard CTC is applied to calculate the loss and yield the final results. These three parts compose an end-to-end trainable network. The proposed new architecture achieves the highest performances on ORAND-CAR-A and ORAND-CAR-B with recognition rates 89.75\% and 91.14\%, respectively. In addition, the experiments on a generated captcha dataset which has much longer string length show the potential of the proposed network to handle long strings.

\end{abstract}

\section{Introduction}

Recent years, with the advancement of deep learning, handwritten digit string recognition (HDSR) has archived great improvements \cite{Diem2014ICFHR, gattal2017segmentation, Saabni2016Recognizing}. An intuitive approach to recognize these handwriting strings is to segment string images into pieces which correspond to single characters or part of them, then combine the recognition results of these pieces with path-search algorithms to get global optimal results. These methods are known as over-segmentation strategy. Wu et al. \cite{Diem2014ICFHR} transformed the string image into a sequence of primitive image segments after binarization, then combined these segments to generate candidate character patterns, forming a segmentation candidate lattice. After that a beam search algorithm was used to find an optimal path over the candidate lattice. This method won the first place on the ICFHR2014 HDSR competition \cite{Diem2014ICFHR}. Saabni \cite{Saabni2016Recognizing} used sliding window and deep neural network to attain high recognition rates. Gattal et al. \cite{gattal2017segmentation} applied three segmentation methods to handle handwritten digit strings by combining these segmentation methods depending on the configuration link between digits. But this kind of methods faced many problems in practice, such as various handwritten styles, connected characters or background noises.

An alternative to handle such sequence recognition task is segmentation-free methods.
Benefitting from the ability of modelling the alignment between inputs and labels directly, connectionist temporal classification (CTC) \cite{graves2006connectionist} is specifically suitable for temporal classification tasks, such as speech and string recognition. CTC is often used as an output layer for recurrent neural network(RNN). In practice such RNN-CTC framework usually combines with a deep neural network, which generates the feature representation of inputs. Messina \cite{messina2015segmentation} firstly applied a LSTM-RNN model to off-line Chinese handwritten text recognition. Without well-designed architecture it achieved competitive performance with the state-of-the-art of tradition method \cite{wang2012handwritten}. Shi et al. \cite{shi2016end} proposed a network with CNN and RNN named CRNN and applied it to scene text recognition. The CRNN is built with Torch. We rebuild the experimental environment in our machine and apply it to HDSR for comparison.

In this paper, we propose a new network based on RNN-CTC framework. First we use the more efficient Residual network \cite{he2016deep}, which was the champion of ILSVRC 2015 classification task, to extract more discriminative feature sequences. Then we modify the standard bi-direction LSTM by adding fully connected layer before combining the two directions for convergence. At the top of our model, a standard CTC is used to calculate the loss and yield the recognition result. By taking the advantages of these models, this new model works well in HDSR task. Compared with the submitted methods in ICFHR 2014 HDSR competition, as well as CRNN, our approach makes significant improvements and achieves the state-of-the-art performance. We conduct our experiments\footnote{https://github.com/LPAIS/HDSR-with-RNN} with Caffe \cite{jia2014caffe} toolkit.

The rest of this paper is organized as follows. In Section 2 we describe the methods. Then, the details of our experiments are presented in Section 3. Section 4 concludes this paper and discusses the future work.

\section{The Proposed Architecture}

The main idea of our model is using a recurrent neural network to model contextual information, namely, the features extracted by a powerful convolutional neural network from raw images and yield elementary results, then get the final recognition results with the output layer connectionist temporary classification.

\subsection{Feature extractor: Convolutional Neural Networks}

Convolutional neural networks are successful in most computer vision tasks. It remains the space structure of image then fully connected network and generates highly-efficient features that defeats traditional methods. CNN has great improvements since it was put forward. Many fantastic CNN architectures were proposed such as AlexNet \cite{krizhevsky2012imagenet}, GoogleNet \cite{szegedy2015going} and network in network(NIN) \cite{lin2013network}.

%A network with residual learning can be called ResNet.
When the networks going deeper, a degradation problem has been exposed. In order to address this issue, He et al. \cite{he2016deep} introduced a deep residual learning framework, i.e. the ResNet.
The essential structure of ResNet is shortcut connection. Shortcut connections are those that skip one or more layers. With this kind of connections, we can handle the vanishing gradient problem and build deeper networks, which means that we can get more excellent feature representations. In practice the way of shortcut connection is flexible according to specific tasks.

In our model, we design a 10 layers residual network without global pooling layers. With the reason of connecting a deep RNN following, we don't employee much deep CNN to avoid divergence. We take the advantages of the residual learning to enhance gradient propagation.

\subsection{Sequence labelling: Recurrent Neural Network}

A recurrent neural network is a class of neural network models where many connections among its neurons form a directed cycle. With self-connections, it has an important benefit to use contextual information when mapping between input and output sequences. But for traditional RNN, the range of context that can be in practice accessed is quite limited due to vanishing gradient problem. One solution is to impose a memory structure into the RNN, resulting in the so-called long-short time memory (LSTM) \cite{hochreiter1997long} cell. Such LSTM version of RNN is shown to overcome some fundamental problems of traditional RNN and can be able to efficiently learn to solve long time dependency problems. Nowadays, LSTM becomes one of the most widely used RNN.

For sequence labelling task it is beneficial to have access to future as well as past context. However, the standard LSTM only consider the past information and ignore future context. An alternative solution is to add another LSTM to handle data reversely, which is so-called bi-direction LSTM \cite{graves2005framewise}, short for BiLSTM. BiLSTM presents each training sequence forwards and backwards to two separate LSTM layers, both of which are connected to the same output layer. This structure provides the output layer with complete past and future context for every point in the input sequence. %The BilSTM used in our model is shown is Fig .\ref{bilstm}.

\subsection{Connectionist temporal classification}

Connectionist temporal classification \cite{graves2006connectionist} is a kind of output layer. It has two main functions. One is to calculate the loss, the other is to decode the output of RNN.

For a sequence labelling task, the labels are drawn from a set $A$ (in HDSR task, $A$ is the ten digits). With an extra label named \emph{blank}, we get a new set $A^{'} = A \cup \{\emph{blank}\}$, which is used in reality. The input of CTC is a sequence $y = y_1,...,y_T$, where T is the sequence length. The corresponding label donates as $I$ over $A$. Each $y_i$ is a probability distribution on the set $A^{'}$. We define a many-to-one function $\mathscr{F}: A^{'T} \mapsto A^{\leq T}$ to resume the repeated labels and blanks. For example $\mathscr{F}(1-22--333--) = 123$. Then, a conditional probability is defined as the sum of probabilities of all $\pi$ which are mapped by $\mathscr{F}$ onto $I$:

\begin{equation}
 p(I|y) = \sum_{\pi\in \mathscr{F}^{-1}(I)}p(\pi|y)
\label{eq1}
\end{equation}
where the conditional probability of $\pi$ is defined as:

\begin{equation}
p(\pi|y)=\prod_{t=1}^{T}{y_{\pi_{t}}^{t}}
\end{equation}
${{y_{\pi_{t}}^{t}}}$is the probability of having label $\pi_{t}$ at timestep $t$. Directly computing Eq.\ref{eq1} is not feasible. In practice, Eq.\ref{eq1} is usually calculated using the forward-backward algorithm.

Donate the training dataset by $S = ({\bm{X},  \bm{I}})$, where $X$ is the training image and $I$ is the ground truth label sequence. The CTC object function $\mathscr{O}(S)$ is defined as the negative log probability of ground truth  all the training examples in training set $S$,

\begin{equation}
\mathscr{O}(S) = - \sum_{(x, i)\in {S}} \log p(I|y)
\end{equation}
where $y$ is the sequence produced by the recurrent layers from $x$. Therefore, the network can be end-to-end trained on pairs of images and sequences, without the procedure of manually labelling all individual components in training images.

\begin{table}[!ht]
\caption{ Network configuration summary. The first row is the top layer. ¡¯k¡¯, ¡¯s¡¯, ¡¯p¡¯
stand for kernel, stride and padding sizes respectively. The `reverse' layer reverses the input. For example, a input of `reverse' layer is $x_1,...,x_n$, the corresponding output is $x_n,...,x_1$.}
\label{table1}
\begin{center}

%\begin{tabular}{|p{2cm}< {\centering}|c|c|}
\begin{tabular}{|c|c|c|}
\hline
 Type & \multicolumn{2}{c|}{Configuration}    \\

\hline

CTC & \multicolumn{2}{c|}{Calculate the loss and decode }  \\

\hline

Elewise & \multicolumn{2}{c|}{Sum }  \\

\hline

Reverse & {Reverse features} & - \\

\hline

InnerProduct & {\#units:11} & {\#units:11}  \\

\hline

InnerProduct & {\#units:100} & {\#hunits:100}  \\

\hline

LSTM & {\#hidden units:100} & {\#hidden units:100}  \\

\hline

LSTM & {\#hidden units:100} & {\#hidden units:100}  \\

\hline

Reverse & {Reverse features} & - \\

\hline

permuted & \multicolumn{2}{c|}{permute the blob to fit lstm} \\

\hline
Eltwise & \multicolumn{2}{c|}{Sum} \\

\hline

Convolution & \#maps:512, k3x3, s1x1, p1x1 & \#maps:512, k1x1, s2x2\\

\hline

Convolution & \#maps:512, k3x3, s2x2, p1x1 & - \\

\hline

Eltwise & \multicolumn{2}{c|}{sum} \\

\hline

Convolution & \#maps:256, k3x3, s1x1, p1x1 & \#maps:256, k1x1, s2x2\\

\hline

Convolution & \#maps:256, k3x3, s2x2, p1x1 & - \\

\hline

Eltwise & \multicolumn{2}{c|}{Sum} \\

\hline

Convolution & \#maps:128, k3x3, s1x1, p1x1 & \#maps:128, k1x1, s2x2\\

\hline

Convolution & \#maps:128, k3x3, s2x2, p1x1 & - \\

\hline

Eltwise & \multicolumn{2}{c|}{Sum} \\

\hline

Convolution &  \#maps:64, k3x3, s1x1, p1x1  & - \\

\hline

Convolution &  \#maps:64, k3x3, s1x1, p1x1  & - \\

\hline

MaxPooling & \multicolumn{2}{c|}{k3x3, s1x1} \\

\hline

Convolution & \multicolumn{2}{c|}{\#maps:64, k5x5, s1x1, p1x1} \\

\hline

Input & \multicolumn{2}{c|}{ Input raw image} \\

\hline

\end{tabular}
\end{center}
\end{table}

\section{Experiments}

To evaluate the effectiveness of proposed model, we designed two experiments. One is to show the recognition performance on public datasets, the other is to verify the potential of our model for long digit strings recognition.

The network configuration used in our experiments is described in Table \ref{table1}. The input images are resized to fixed size. The CNN part is derived from the residual network with necessary modifies. We reduce the kernel size to better suit for CTC decoding and remove the global pooling layer in ordinary ResNet. After deep convolutional layers, there is a bidirectional LSTM, each direction has two layers of LSTM.

For implementation, by using the fundamental LSTM layer in Caffe and a custom reverse layer, we build the Bi-LSTM layer in C++ without combining them into a single Bi-LSTM layer. This proposed architecture contains deep convolutional layers and deep recurrent layers which are known to be hard to train. In practice the network is trained with ADADELTA, setting the essential parameter delta to $10^{-6}$.

Our experiments are performed on a DELL workstation. The CPU is Intel Xeon E5-1650 with 3.5GHz and the GPU is NVIDIA TITAN X. The software is the latest version of Caffe \cite{jia2014caffe} with cuDNN V5 accelerated on Ubuntu 14.04 LTS system. The average testing time is 3.5ms per image.

We use a hard metric to evaluate our methods. We calculate the recognition rate, which is defined as the number of correctly recognized digit strings divided by the total number of strings. Because there are more than one digit label in a string, we consider one string being recognized correctly only if when all labels are recognized correctly.

\begin{figure*}[!htp]
\begin{center}
\subfigure[CVL HDS]{
\centering
\includegraphics[width=0.8\columnwidth]{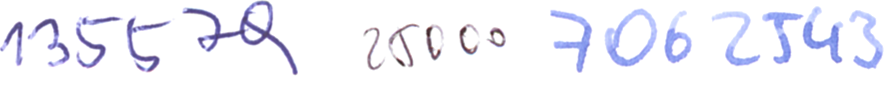}}
\subfigure[CAR-A]{
\includegraphics[width=0.8\columnwidth]{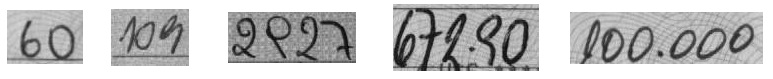}}
\subfigure[CAR-B]{
\includegraphics[width=0.8\columnwidth]{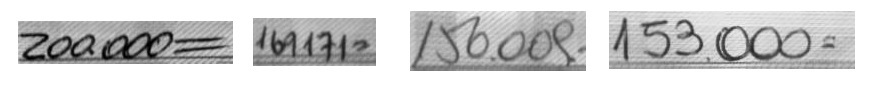}}
\subfigure[G-Captcha]{
\includegraphics[width=0.8\columnwidth]{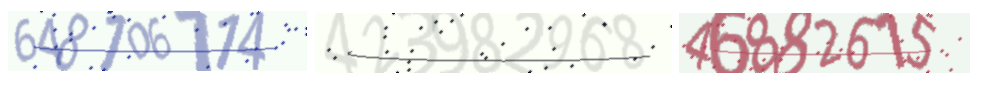}}
\caption{Samples of the datasets used in our experiments.}
\label{sampleimg}
\end{center}
\end{figure*}

\subsection{Datasets}

There are two public datasets used in our experiments. The first dataset, named as Computer Vision Lab Handwritten Digit String (CVL HDS), is collected from about 300 writers. The variability of writers brings high variability with respect to handwritten styles. The CVL HDS dataset has 7960 images, from which 1262 images for training and the other 6698 images for testing. Some examples are shown in Figure \ref{sampleimg}(a) with different written styles.

\begin{table}[!ht]
\caption{  Distribution of the databases with respect to string length. }
\label{string_length}
\begin{center}

\begin{tabular}{|c|c|c|c|c|c|c|c|c|}
\hline
 & \multicolumn{4}{c|}{training set}  & \multicolumn{4}{c|}{testing set}    \\
 \hline
 len & CVL & CAR-A &  CAR-B & G-Captcha & CVL & CAR-A & CAR-B & G-Captcha    \\
 \hline
 below 7  & 883 & 1978 & 2862 & 0 & 4933 & 3686 & 2767 & 0      \\
\hline
     7 & 379 & 29 & 137  & 0 & 1765 & 87 & 157 & 0  \\
\hline
     8 &  0 &2  & 1 & 1500     &  0 &11 & 2 & 2000\\
\hline
     9 &  0 &0  & 0 & 1500     &  0 &0  & 0 & 2000\\
\hline
     10 & 0 & 0 & 0 & 1500     &  0 &0  & 0 & 2000\\
\hline
     11 & 0 & 0 & 0 & 1500     &  0 &0  & 0 & 2000\\
\hline

\end{tabular}
\end{center}
\end{table}

The other dataset is ORAND-CAR, consisting of 11719 images obtained from the Courtesy Amount Recognition (CAR) field of real bank checks. The ORAND-CAR images come from two different sources with different characteristics. Considering the two different sources, ORAND-CAR is divided into two subsets, ORAND-CAR-A and ORAND-CAR-B, which are abbreviated to CAR-A and CAR-B.

The CAR-A database consists of 2009 images for training and 3784 images for testing. The CAR-B database consists of 3000 training images and 2926 testing images. Some samples are shown in Figure \ref{sampleimg}(b)-(c).

String lengths of samples in these two datasets are mostly not larger than 7. For string recognition, the longer strings are the harder task is. So we create two captcha datasets by using a Python package named `captcha'\footnote{https://pypi.python.org/pypi/captcha/0.1.1}. This package can generate arbitrary length captcha images with dirty background, and the styles of digits are varieties, which are similar to human handwriting.

The generated captcha dataset is named G-Captcha, in which string length is extended to 11. The distribution of the four different datasets with respect to string length are show in table. \ref{string_length}. With the Python package `captcha', we can create arbitrary lengths of captcha images. This dataset contains 14,000 images, in which 6,000 for training and 8,000 for testing. The examples of G-Captcha are shown in Figure.\ref{sampleimg}(d).

\subsection{Results and analysis}

The experimental results are shown in table \ref{table2}. We can see that the proposed network achieves the state-of-the-art on both ORAND-CAR-A and ORAND-CAR-B with a huge advance. But it performs very bad on CVL HDS dataset. On the other side, traditional methods performed outstandingly. There are 300 writers that contribute to CVL HDS. For each writer, 26 different digit strings were collected. Only 10 kinds of strings occur in training set. For segmentation methods, this is not a problem because the total categories of numbers are ten. But for methods based on RNN-CTC, it gets into trouble due to the lack of sample diversity. The CRNN architecture \cite{shi2016end} which is also derived from RNN-CTC faces the same problem. The result on G-Captcha shows the strong ability of handling very long strings.

\begin{table}
\caption{ Recognition rates of different models on the datasets described above. (The top five methods are proposed on the HDSRC 2014\cite{Diem2014ICFHR}, the following two are proposed in newest papers. Especially, last but one method uses the ORAND-CAR dataset as a whole.)}
\label{table2}
\begin{center}

\begin{tabular}{|p{2.5cm}< {\centering}|p{2cm}< {\centering}|p{2cm }< {\centering}|p{2cm }< {\centering}|p{2cm }< {\centering}|}
\hline
Methods & CAR-A & CAR-B & CVL HDS  & G-Captcha  \\
 \hline
Tebessa I \cite{Diem2014ICFHR} &0.3705 & 0.2662 & 0.5930 & - \\

\hline
Tebessa II \cite{Diem2014ICFHR} & 0.3972 & 0.2772 & 0.6123 & -\\
\hline
Singapore \cite{Diem2014ICFHR} & 0.5230 & 0.5930 & 0.5040 & - \\
\hline
Pernambuco \cite{Diem2014ICFHR} & 0.7830 & 0.7543 & 0.5860  & - \\

\hline
BeiJing \cite{Diem2014ICFHR} & 0.8073 &  0.7013 & \bf{0.8529} & - \\

%\hline
%FSPP\cite{wang2017sequence} & 0.8261 &  0.8332 & 0.7923& -  \\

\hline
CRNN \cite{shi2016end} & 0.8801 &  0.8979 & 0.2601& 0.9312  \\

\hline
Saabni \cite{Saabni2016Recognizing} & \multicolumn{2}{c|}{0.8580}   & -& -  \\

\hline
Proposed & \bf{0.8975} & \bf{0.9114} & 0.2707 & \bf{0.9515} \\

\hline

\end{tabular}
\end{center}
\end{table}

 The above experiments demonstrate that the proposed method can work well on handwritten digit string recognition. Apart from the shorter length strings in public datasets, it can recognize much longer strings. Although this method fails to handle CVL HDS, it is also an efficient approach to deal with digit string recognition, even if the string is long.

\section{Conclusion and Future work}

In this paper, we have presented a new model based on RNN-CTC architecture. Combined with ResNet, we archive the new arts on the handwritten digit string recognition benchmarks ORAND-CAR with a great improvement. The experiments on G-Capthca dataset, which consists of much longer string images, indicate our model is very competitive of handling long sequence labelling. But it gets into trouble when handling data like CVL HDS. It seems a common failing for such kind of method. In future, we will refine our model based on elaborate analysis the results in this paper, and exploit the potential on relative areas.

\end{document}